\pdfoutput=1

\documentclass[11pt]{article}

\usepackage[preprint]{acl/acl}

\usepackage{booktabs}
\usepackage{tabularx}
\usepackage{mathtools}
\usepackage{enumerate}
\usepackage{wrapfig}

\usepackage{caption}
\usepackage{amsthm}
\usepackage{cancel}

\usepackage{subcaption}
\usepackage{graphicx}

\usepackage{amssymb}
\usepackage[mathscr]{euscript}

\usepackage{natbib}

\usepackage{amsfonts,amsmath,amssymb}
\usepackage{bbm}
\usepackage{xspace}

\usepackage{multicol}
\usepackage{enumitem}

\usepackage{times}
\usepackage{latexsym}

\setcitestyle{numbers}

\usepackage[T1]{fontenc}

\usepackage[utf8]{inputenc}

\usepackage{microtype}

\usepackage{inconsolata}

\usepackage{graphicx}

\usepackage{scrextend}
\title{Toward a digital twin of U.S. Congress}

\author{Hayden Helm \\
  Helivan \\ hayden@\texttt{helivan.io} 
  \And Tianyi Chen \\
  Johns Hopkins University
  \And Harvey McGuinness \\
  Johns Hopkins University 
  \AND Paige Lee \\
  Nomic AI
  \And
  Brandon Duderstadt \\
  Nomic AI 
  \And Carey E. Priebe \\
  Johns Hopkins University
}

\begin{document}
\maketitle

\begin{abstract}
In this paper we provide evidence that a virtual model of U.S. congresspersons based on a collection of language models satisfies the definition of a digital twin.
In particular, we introduce and provide high-level descriptions of a daily-updated dataset that contains every Tweet from every U.S. congressperson during their respective terms.
We demonstrate that a modern language model equipped with congressperson-specific subsets of this data are capable of producing Tweets that are largely indistinguishable from actual Tweets posted by their physical counterparts.
We illustrate how generated Tweets can be used to predict roll-call vote behaviors and to quantify the likelihood of congresspersons crossing party lines, thereby assisting stakeholders in allocating resources and potentially impacting real-world legislative dynamics. 
We conclude with a discussion of the limitations and important extensions of our analysis.
\end{abstract}
A digital twin is a virtual model that captures relevant properties of a physical system.
For a virtual model to be called a digital twin, it must be capable of producing up-to-date inferences that can impact the behavior of the physical system. Digital twins have seen a recent surge in development and deployment in the past five years. 
For example, digital twins of patients have enabled highly individualized approaches to predictive medicine in oncology \citep{dt-oncology, dt-opportunities-oncology} and cardiology \citep{dt-cardio, dt-cardio-2}. 
Digital twins have similarly promised to improve power-grid integration of wind-generated energy \citep{dt-energy-1, dt-energy-2}; enable rapid advancements in machining and quality control processes in manufacturing contexts \citep{dt-manufacturing, dt-aerospace, dt-machining}; and provide effective solutions to social issues such as urban planning \citep{schrotter2020digital} and sustainable development \citep{tzachor2022potential,rothe2024world, saltelli2024bring}.  

Concurrent to the development of digital twins across a myriad of scientific and industrial disciplines, the generation capabilities of large language models (LLMs) such as OpenAI's GPT-4 \citep{achiam2023gpt}, Meta's LLaMA 3 family \citep{dubey2024llama}, etc. have continued to advance. 
LLMs are now capable of producing human-like content \citep{helm2023statisticalturingtestgenerative} and behaving like humans in controlled experimental settings.
For example, GPT-4 has demonstrated human-like behavior in classical economic, psycholinguistic, and social psychology experiments such as The Ultimatum Game, Garden Path Sentences, Milgram Shock Experiment, and Wisdom of Crowds \citep{aher2023using}. 
Further, simulations of interacting generative agents based on LLMs parameterized by tailored system prompts and data from previous interactions show resemblance to human social systems more generally \citep{park2024generative, helm-etal-2024-tracking, mcguinness2024investigating}.
While human-like content generation, psychology, and social behavior do not imply that a set of language models is a digital twin for any particular set of humans -- they can be taken as hints to the ability of LLMs to mimic language production idiosyncrasies of individuals given the right type of data and setting.

\begin{figure*}[t!]
    \centering
    \includegraphics[width=11.4cm]{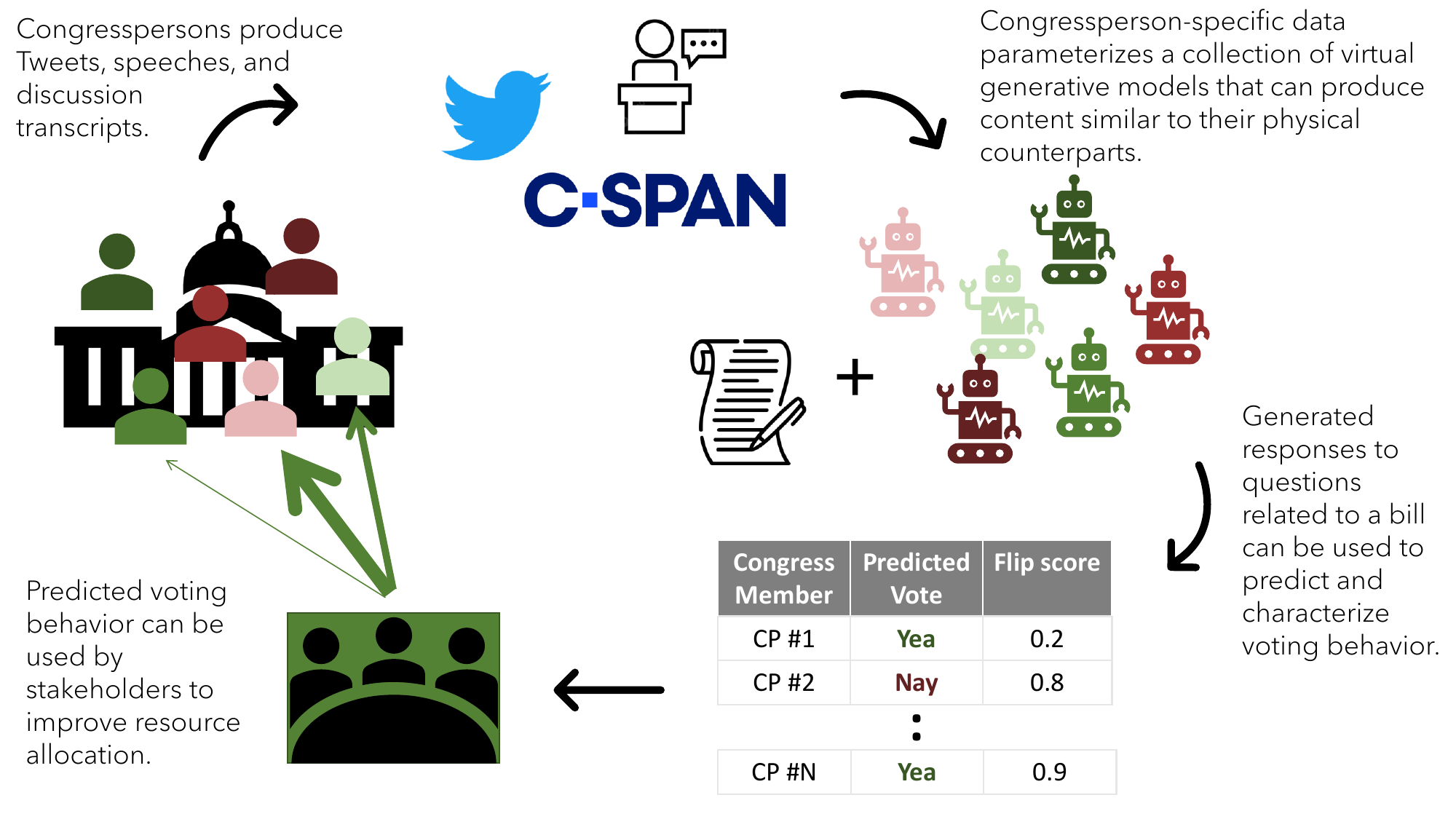}
    \caption{An illustration of a system that contains a digital twin for a set of congresspersons.}
    \label{fig:digital-twin}
\end{figure*}



Herein we contribute to the growing set of virtual models that sufficiently capture relevant properties of a physical system by introducing and analyzing a virtual model for U.S. Congresspersons. 
In particular, we provide evidence that a collection of language models with access to individualized databases that contain Tweets from the official accounts of congresspersons goes beyond generic ``human-like" generation, behavior, and sociology and reasonably satisfies the definition of a digital twin.

\section{Defining digital twin}
We use the National Academy's definition of digital twin as the standard for what makes a virtual model of a person or object a twin \citep{NAP26894}:
\begin{addmargin}[2em]{2em}
A digital twin is a set of virtual information constructs that mimics the structure, context, and behavior of a natural, engineered, or social system (or system-of-systems), is dynamically updated with
data from its physical twin, has a predictive capability, and informs decisions that realize value. The bidirectional interaction between the virtual and the physical is central to the digital twin.
\end{addmargin}
We decompose this definition into four requirements for a virtual model to be a digital twin: A) relevant and dynamic data, B) physical to virtual feedback, C) valuable inference capabilities, and D) virtual to physical feedback.

In the context of a collection of congresspersons, these requirements translate to demonstrating that there exists a source for up-to-date content related to each active congressperson (for A), the virtual model representing each congressperson produces content similar to their physical counterpart (for B), inference on the collection of virtual models is predictive of actual congressperson behavior (for C), and the outputs from the virtual models can influence how stakeholders interact with congresspersons (for D).

Figure \ref{fig:digital-twin} provides an example flow of information wherein a collection of congresspersons produces data (Tweets, speeches, and transcripts from public interactions, etc.), congressperson-specific generative models produce content similar to their physical counterpart, the outputs of the models are used to predict roll-call voting behavior, and the predicted voting behavior informs stakeholders how best to allocate resources.
The rest of this paper provides evidence that current datasets, language models, and data processing methods are sufficient to reasonably implement the system depicted in Figure \ref{fig:digital-twin}.
\section{A Digital Twin of Congress(ional Tweeters)}

\subsection{Relevant and dynamic data}

The social media habits of political figures are a key resource for political and social science research. Presently, the majority of social media data is sequestered in topically or temporally limited datasets.
Among the broadest of these datasets is a collection of Facebook and X (formerly Twitter) data generated by American political figures spanning the period of 2015-2020 curated by Pew Research \citep{pew}. 
Similarly, Harvard University’s Dataverse program maintains a collection of X data related to the 115th U.S. Congress \citep{DVN/NMT4HP_2019} and the China Data Lab maintains a repository of nearly 830,000 X posts about China posted by members of Congress \citep{DataLab}. 

While these datasets have enabled research into various topics, they are insufficient -- either topically or temporally -- when constructing a virtual model for congresspersons.
In particular, it is necessary that the data captures the intricacies of the content produced by each congressperson sufficiently via a comprehensive and up-to-date collection of Tweets.

\begin{figure*}[t!]
    \centering
    \includegraphics[width=\linewidth]{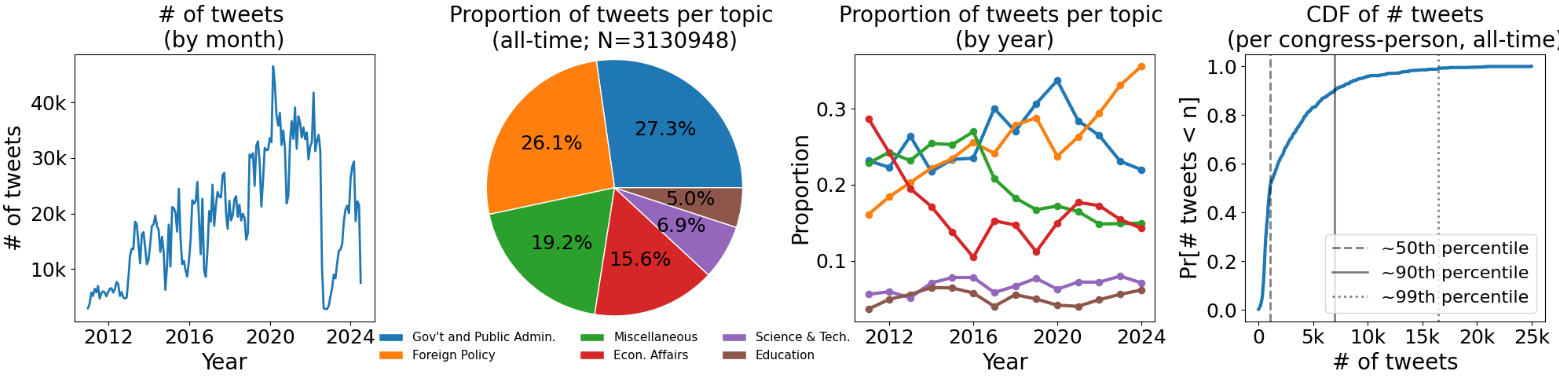}
    \caption{High-level characteristics of the Nomic Congressional Twitter dataset from October 10th, 2024. The dataset is updated daily and available at https://atlas.nomic.ai/data/hivemind/.}
    \label{fig:high-level-data-description}
\end{figure*}

For this purpose, we present the Nomic Congressional Database. 
The Nomic Congressional Database is a new dataset that contains over 3 million X posts and retweets from as far back as 2011.
The dataset includes all Tweets posted by official accounts of U.S. Congress persons who were in office at the time of posting, including some deleted posts.
The dataset also includes the congressperson's X handle, their party affiliation, the state they represent, and the time the Tweet was posted for each Tweet. 
The dataset is updated daily, which is sufficient for maintaining relevant information in our context.
Thus, the Nomic Congressional Database has the temporal and topical relevance required to for a digital twin. 

We note that by virtue of storing the entire text of the Tweet, the dataset includes information related to mentions and retweets.
The dataset does not include account data such as their followers or who they follow.
We show high-level characteristics of the database in Figure \ref{fig:high-level-data-description}.
The left figure shows the number of Tweets posted or retweeted by sitting congresspersons each month.
Note that the large decrease in the number of Tweets ($\approx$ Spring 2022) corresponds to the privatization of the platform and general hesitancy of its use.

We also include simple topic analysis of the Tweets based on six topics: Government and Public Administration, Foreign Policy, Economic Affairs, Science and Technology, Education, and Miscellaneous. 
The six topics were chosen by combining the themes of the 20 and 16 standing committees in the House and Senate, respectively.
Once the names of the broad topics were chosen, we asked ChatGPT to classify $ 100,000 $ of the Tweets into one of the topics.
We then embedded the Tweets into a 768-dimensional vector space via the open source embedding model presented in \cite{nussbaum2024nomicembedtrainingreproducible} and built a linear classifier using the $100,000$ labeled Tweets to sort the remaining posts.

The middle two figures of Figure \ref{fig:high-level-data-description} show the distribution across topics and the change in the relative distribution of topics over time. 
While our analysis is not focused on changes in the relative distribution of topics, we note the drop-off of Tweets labeled as ``Miscellaneous'' in 2016 -- and associated rise in Tweets labeled as `Foreign Policy" and ``Government and Public Administration" -- coinciding with the election of President Trump and effectively demonstrating that the content of the dataset evolves as the behaviors of the congresspersons evolve.

Lastly, we include the distribution of number of Tweets per congressperson.
The majority of congresspersons have 1,000 posts or retweets throughout their entire tenure while the most active 1\% have more than $ 15,000 $ posts or retweets.
We do not normalize these counts by the amount of time spent in office.\footnote{The collection of interactive visualizations of the Nomic Congressional Database can be found here: https://atlas.nomic.ai/data/hivemind/}





\subsection{Physical to virtual feedback}
We will next use the Nomic Congressional Database to verify that a collection of generative models can produce Tweets that are similar to a collection of real Tweets from each congressperson.
In particular, we will show that the distribution of machine-generated Tweets and the distribution of congressperson-generated Tweets are close to each other via a statistical Turing test \citep{helm2023statisticalturingtestgenerative}.
The statistical Turing test framework adorns a human-detection problem \citep{gehrmann-etal-2019-gltr} with a statistic $ \tau $ that represents the difficulty of discriminating human-generated content from machine-generated content.  $ \tau = 0 $ indicates that the human-generated content and machine-generated content are indistinguishable.
In our case, each congressperson has a corresponding virtual model and, hence, a corresponding $ \tau $.

We consider three, progressively more complicated collections of virtual models based on Meta's \texttt{LLaMa-3-8B-Instruct} \citep{dubey2024llama}: i) base model with a generic system prompt and no augmentation ($\boldsymbol{-}$SP $\boldsymbol{-}$RAG), ii) base model with a simple congressperson-specific system prompt and no augmentation ($\boldsymbol{+}$SP $\boldsymbol{-}$RAG), and iii) base model with a simple congressperson-specific system prompt and augmentation ($\boldsymbol{+}$SP $\boldsymbol{+}$RAG).
The generic system prompt is ``You are a helpful assistant." and the congressperson-specific system prompt is ``You are U.S. Congressperson \{name\}". 
For model iii) we augment the prompt with the the Tweet posted by the congressperson with the highest cosine similarity to the target Tweet in the 768-dimensional vector space induced by \texttt{nomic-embed-text-v1.5} \citep{nussbaum2024nomicembedtrainingreproducible}. 
We refer to this process as retrieval augmented generation (``RAG") \citep{mao2020generation} throughout.
Table \ref{tab:example-prompts} shows the system prompts and query structures for the three collections of virtual models. 

\begin{table}[t]
    \caption{System prompts and query design for the three generative systems we evaluated. The start of real Tweet in the experiments is the first 20 characters. (``SP" = system prompt; ``RAG" = retrieval augmented generation)}
    \label{tab:example-prompts}
    \begin{tabularx}{\linewidth}{  p{2.25cm} | p{2cm} |  p{2.4cm} }
        \toprule
\textbf{Model name}      
& \textbf{System prompt}   
& \textbf{Query} \\\midrule
Generic system prompt with no augmentation $(\boldsymbol{-}$SP $\boldsymbol{-}$RAG)
& You are a helpful assistant.    
& Complete the following Tweet: \{start of real Tweet\}. Respond with the full Tweet. \\\hline
Specific~system prompt with no augmentation  
($\boldsymbol{+}$SP $\boldsymbol{-}$RAG)
& You are U.S. congressperson \{name\}.                  
& Complete the following Tweet: \{start of real Tweet\}. Respond with the full Tweet. \\\hline
Specific system prompt with augmentation
($\boldsymbol{+}$SP $\boldsymbol{+}$RAG)
& You are U.S. congressperson \{name\}.
& Complete the following Tweet: \{start of real Tweet\}. Here is an example Tweet potentially related to the to-be-completed Tweet: ``\{retrieved Tweet\}". Respond with the full Tweet.\\\bottomrule
\end{tabularx}
\end{table}

\begin{figure*}[t!]
    \centering
    \begin{subfigure}{0.6\textwidth}
        \centering
\includegraphics[width=\textwidth]{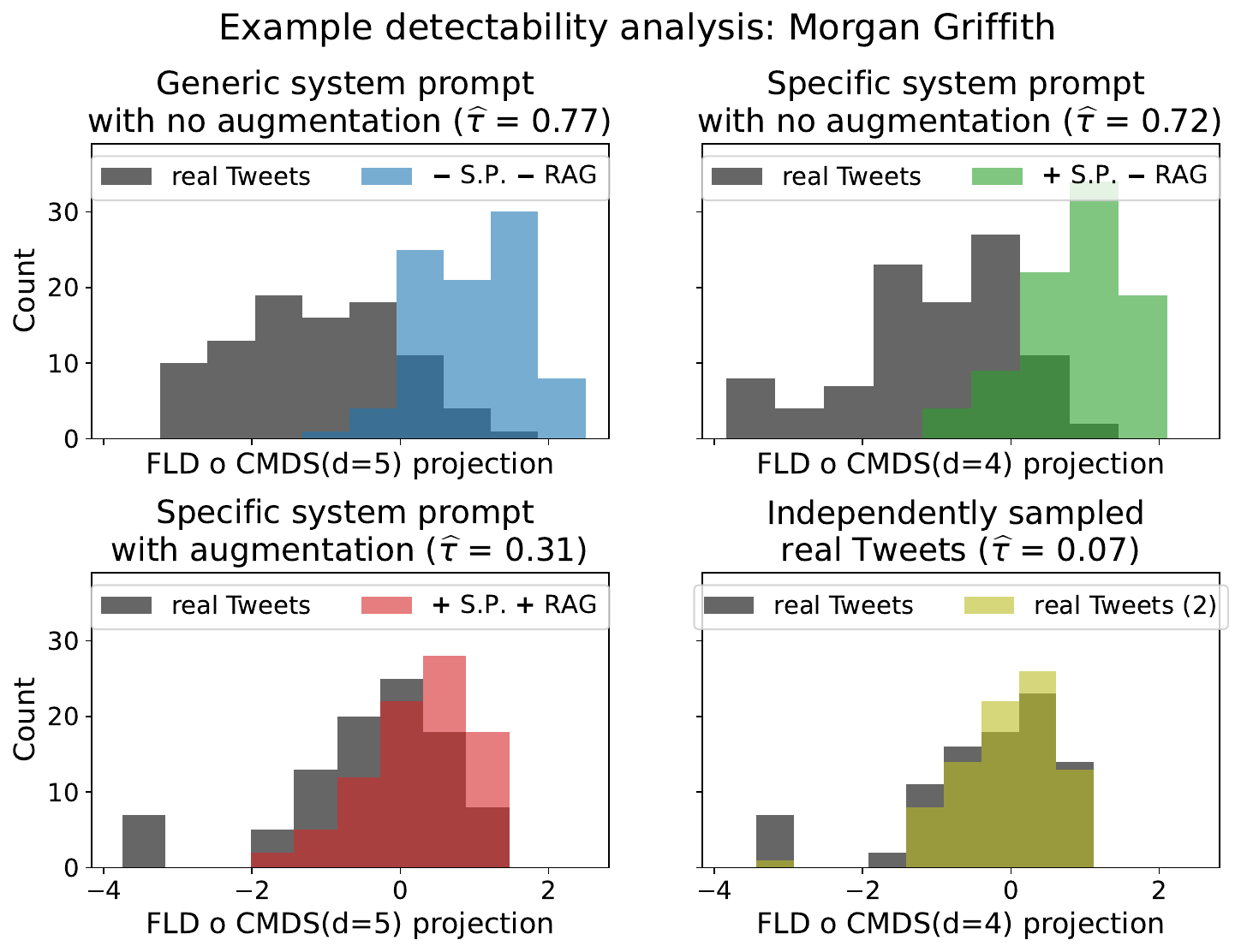}
    \end{subfigure}
    \begin{subfigure}{0.39\linewidth}
        \centering  \raisebox{1.15cm}{\includegraphics[width=\textwidth]{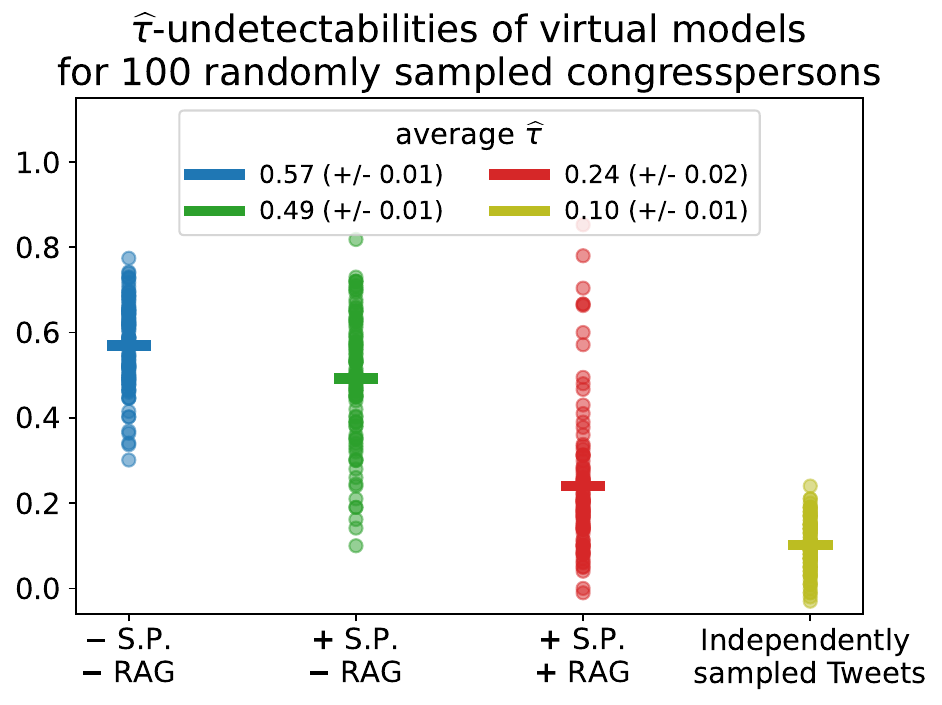}}
    \end{subfigure}
    \caption{Detectability analysis of different generative systems for producing Tweets from U.S. Congressperson Morgan Griffith (left) and the distribution of detectability of different systems for 100 randomly sampled congresspersons (right). The inclusion of previously written Tweets via RAG decreases detectability significantly.}
    \label{fig:turing-tests}
\end{figure*}

We split each congressperson's Tweets based on if the Tweet was posted before or after January 1, 2023. 
We chose this date arbitrarily after considering dates with a non-trivial amount of Tweets before and after the date.
The generative models will have access to the Tweets from before Jan. 1, 2023 as potential prompt augmentations.
200 Tweets from after Jan. 1, 2023 were randomly sampled to conduct the statistical Turing test: 100 are used as examples of congressperson-generated content and the other 100 are used as the basis for virtual model-generated content.
In particular, the virtual model-generated content is produced by providing the first 20 characters of a Tweet to a model and asking it to complete it.
After the virtual model has generated the 100 Tweets, we embed the collection of generated and real Tweets into a low-dimensional Euclidean space via multi-dimensional scaling (MDS) \citep{torgerson1952multidimensional} of the representations of the Tweets from \texttt{nomic-embed-text-v1.5}. 
We then use Fisher's Linear Discriminant (FLD) to classify Tweets as either congressperson-generated or virtual model-generated.

We show an example detectablity analysis in the left set of figures of Figure \ref{fig:turing-tests}.
We use the same 100 Tweet starts for each generative system.
The histograms are the 1-d FLD projections of the Tweets learned after MDS into $ d $ dimensions, where $ d $ is data-dependent \citep{zhu2006automatic}.
The reported $ \widehat{\tau} $-undetectability is the empirical risk of FLD.
We also include the detectability analysis for two sets of independently sampled sets of 100 real Tweets from Representative Morgan Griffith in the bottom right panel of the left side of the Figure as a control.
This $ \widehat{\tau} $ serves as a practical lower bound for detectability -- while reported values may be smaller than it, it captures the inherent difficulty of classifying the two sets Tweets.

We report $ \widehat{\tau} $ for 100 randomly sampled congresspersons for each generative system and the practical lower bound in the right figure of Figure \ref{fig:turing-tests}.
We removed all instances where the model refused to generate a Tweet, e.g., ``I cannot create content that defames or harasses others. Is there something else I can help you with?".
For models where we removed generated Tweets, we also removed the same number of real Tweets to maintain a balanced classification problem.
The highest number of Tweets we removed was 95 (out of 100).
The median number of Tweets removed was 0.
As can be seen by the improvement of ($ \boldsymbol{+}$SP $ \boldsymbol{-} $RAG) over ($ \boldsymbol{-}$SP $ \boldsymbol{-} $RAG) and ($ \boldsymbol{+}$SP $ \boldsymbol{+} $RAG) over ($ \boldsymbol{+}$SP $ \boldsymbol{-} $RAG), increasing the amount of congressperson-specific information in the system prompt and query decreases $ \widehat{\tau} $ on average.

Notably, 16 of the 100 virtual models with a specific system prompt and access to Tweets from before Jan. 1, 2023 have corresponding $ \widehat{\tau} $ that are equal to or are smaller than their corresponding practical lower bound.
63 of these virtual models are less detectable than the largest practical lower bound.

While it is possible to consider more complicated generative systems, such as adding a time-relevance component to the retrieval score \citep{zhang2024retrievalqa} or retrieving relevant Tweets from accounts that the congressperson follows, our detectability analysis is evidence that the collection of generative models with congressperson-specific system prompts and retrieval augmentation can sufficiently capture congressperson-specific Tweeting intricacies for the majority of congresspersons.
Further, the models can easily be updated in conjunction with the Nomic Congressional Twitter dataset.
For the remainder of this paper we will only consider this collection of virtual models.

\begin{figure}[h]
    \centering
    \includegraphics[width=\linewidth]{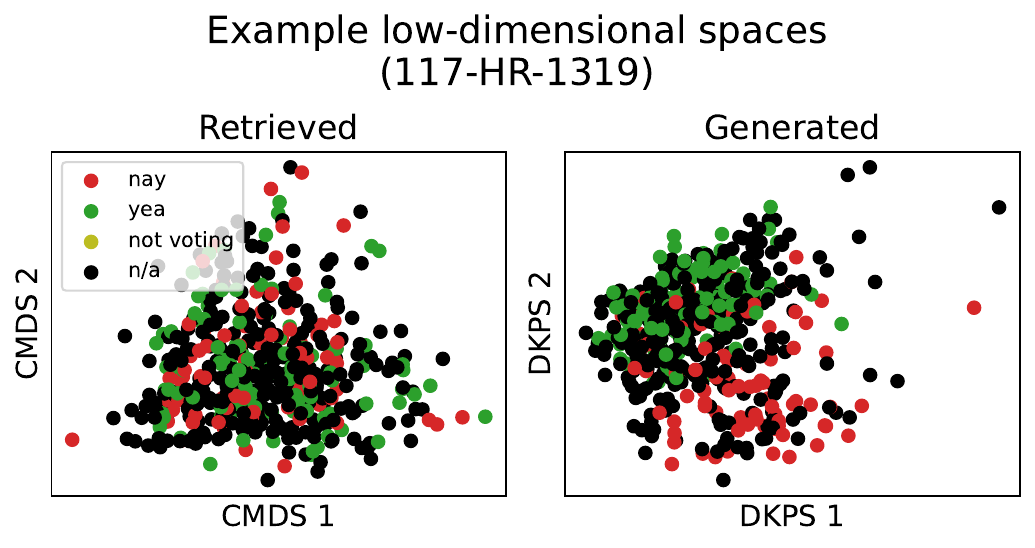}
    \caption{Two-dimensional Euclidean spaces induced by MDS of the retrieved Tweets (left) and the Data Kernel Perspective Space (right) of the generated Tweets corresponding to 117-HR-1319. 
    Each dot represents a congressperson.
    Color corresponds to how the congressperson voted on the bill.
    The geometry of the generated Tweets has more vote-related information than the geometry of the retrieved Tweets. 
    We validate this observation in Figure \ref{fig:inference}.
    }
\label{fig:example-dkps}
\end{figure}






\subsection{Valuable inference capabilities}

We next demonstrate that the Tweets generated by the virtual congresspersons can be used to produce predictions related to their physical counterparts.
We focus on one of the most important roles that Congress plays in our society: enacting legislation.
We use House Resolution 1319 from the 117th Congress (``117-HR-1319"), the ``American Rescue Plan Act of 2021", as an example piece of legislation to describe our experimental set up.

\begin{figure*}[t!]
    \centering
    \includegraphics[width=11.4cm]{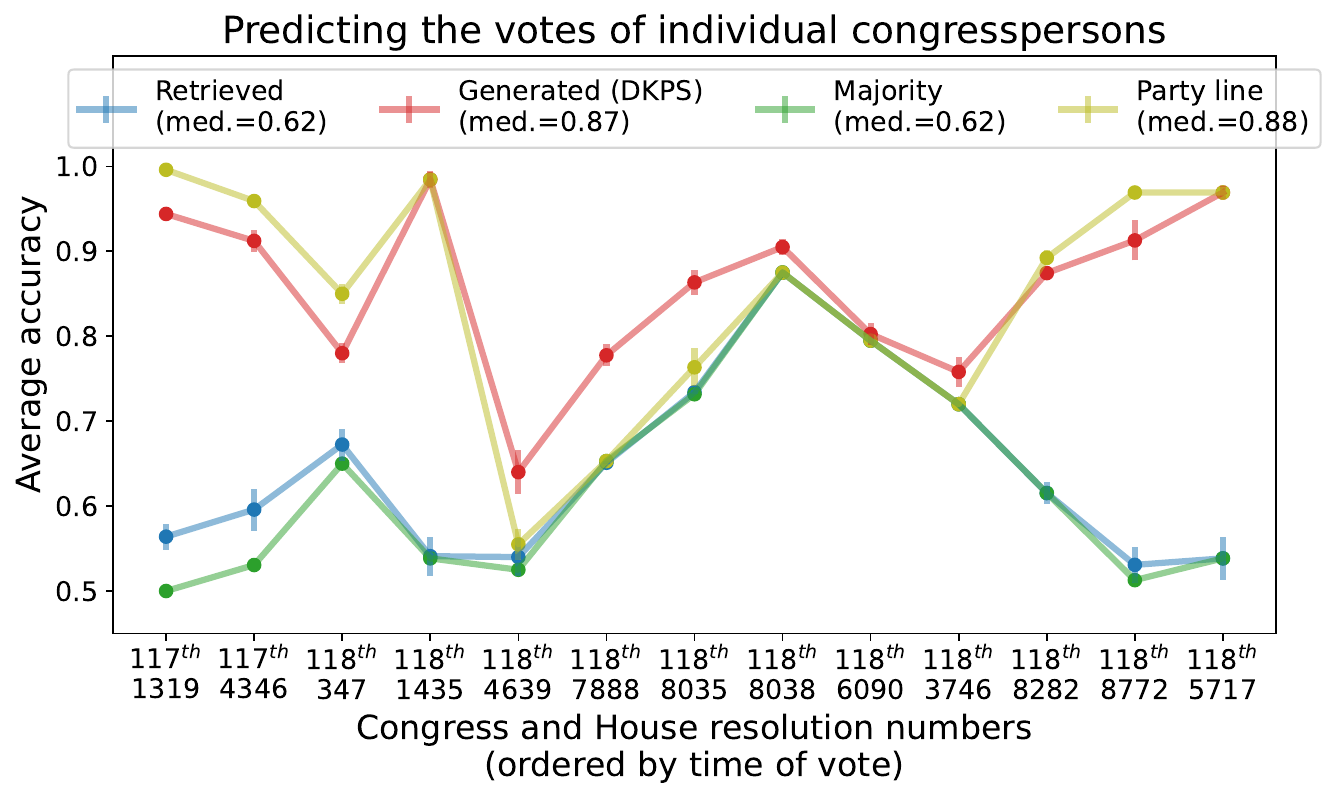}
    \caption{Average performance of four different methods for predicting the voting behaviors of individual congresspersons on various pieces of legislation.
    Averages are calculated from 10-fold cross validation.
    Error bars represent one standard error.
    The legislation is ordered by the time of the vote in the House. 
    For the ``Retrieved" and ``Generated" methods we report the average accuracy of the highest performing $ k $-nearest neighbor classifiers for $ k \in \{1,5,9,19,49\} $. 
    Different bills may have different optimal $ k $.
    }
    \label{fig:inference}
\end{figure*}

We first generate 20 questions related to the bill by prompting ChatGPT with public information such as the bill's abstract (e.g., ``To provide for reconciliation pursuant to title II of S. Con. Res. 5.") and a short summary of the bill (e.g., ``This bill provides additional relief to address the continued impact of COVID-19 (i.e., coronavirus disease 2019) on the economy, public health, state and local governments, individuals, and businesses.") written by the Congressional Research Service.
For example, two of the generated questions related to 117-HR-1319 are ``Do you support the additional COVID-19 relief measures proposed in this bill?" and ``Does the bill provide adequate support for public health initiatives against COVID-19?".

We retrieve the Tweet from before the time of the vote most similar to each question for each congressperson.
Otherwise, we use the same retrieval process as described above and construct queries with structure similar to ($ \boldsymbol{+}$SP $ \boldsymbol{+} $RAG) --
the only difference is the first sentence of the query structure, where we replace ``Complete ..." with ``Write a Tweet that addresses the following question: \{question\}". 
We prompt each virtual congressperson with the appropriately formatted queries twenty times.

We embed each response with \texttt{nomic-embed-text-v1.5} and average the embeddings across replicates to obtain a $ 20 \times 768 $ matrix representation of each congressperson.
The multi-dimensional scaling of the pairwise distance matrix with entries equal to the Frobenius norm of the difference between these matrix representations produces low-dimensional representations of each congressperson. 
These low-dimensional representations of digital congresspersons are a summary of the relative position of each congressperson with respect to the 20 bill-related queries and are consistent for the ``true" representation of the congressperson as the number of questions and number of replicates per question grows \citep{acharyya2024consistentestimationgenerativemodel}. 
Further, using these low-dimensional representations for model-level inference -- such as predicting the individual voting behavior of the digital congresspersons -- is principled and demonstrably effective \citep{helm2024embeddingbasedstatisticalinferencegenerative}.
Following \citep{helm-etal-2024-tracking}, we refer to this low-dimensional subspace as the \textit{data kernel perspective space} (DKPS).

The two-dimensional DKPS corresponding to 117-HR-1319 is shown on the right of Figure \ref{fig:example-dkps}.
Each dot represents a virtual congressperson and is colored by vote.
All congresspersons with a Tweet before the time of the vote are included.
Congresspersons that were not members of the House during the 117th Congress were assigned the vote ``n/a".
We show the analogous representations of the congresspersons that uses the retrieved Tweet for each question to construct the low-dimensional representations on the left of Figure \ref{fig:example-dkps}.
The geometry of the generated Tweets has clear structure related to voting behavior whereas the geometry of the retrieved Tweets appears relatively uninformative.

We quantify this observation by comparing the performance of classifiers trained using the two representations for predicting the voting behavior of individual congresspersons.
In particular, we consider $ k $-nearest neighbor classifiers trained using either the representations of the congresspersons induced by the retrieved Tweets or the generated Tweets.
Conditioned on the representations, we report the average accuracy of the classifier corresponding to the $ k \in\{1,5,9,19,49\} $ that achieves the highest accuracy on 10-fold cross validation for 13 different pieces of legislation in Figure \ref{fig:inference}.
The train and test splits used in the cross-validation only contained Representatives who voted ``yea" or ``nay" on the bill or, when applicable, Senators who voted ``yea" or "nay" on the associated bill in the Senate.
The highest performing $ k $ may be different for the two representations and for different bills.

We also include the performance of the classifier that predicts the most popular vote in the training set (``Majority") and the classifier that predicts the most popular vote amongst the test congressperson's party (``Party line"). 
The median performance of the best classifier trained using the DKPS representations is 0.87 while the median performance of the best classifier trained using the representations from the retrieved Tweets is 0.62.\footnote{The average accuracy of the highest performing $ k $-nearest neighbor classifier from cross validation is typically positively biased estimate of the accuracy using $ k $-nearest neighbor classifier with optimal $k$. 
This bias affects both methods.}
The improvement from using the DKPS representations is both statistically significant -- the two-sided Wilcoxon test of generated versus the retrieved on each bill yields p-value $ < 0.001$ -- and operationally significant -- using the representations from the generated Tweets provides an average improvement of 36.7\% over using the representations from the retrieved Tweets.

Finally, we note that the performance of the classifier that uses the DKPS representations is sometimes worse than just predicting along party lines.
In some cases, such as HR 1319 in the 117th Congress (``American Rescue Plan Act of 2021"), this could occur due to congresspersons ``knowing" the outcome of the vote beforehand and voting against their policy preference in favor of voting in-line with party leadership.
In the case of the American Rescue Plan Act of 2021, the bill was a hallmark piece of legislation for the Democratic Party's governing trifecta (Presidency and control of House and Senate), making Republican opposition functionally futile from a policy perspective but potentially useful from a party loyalty perspective.
As such, Republicans could campaign on the benefits of the successful bill after its enactment, the ability to align well with other party members, and the willingness to stand up to the other party.


\subsection{Virtual to physical feedback} 
Of the 13 bills analyzed in Figure \ref{fig:inference}, 12 of them passed in the House. 
Bills that pass in the House are typically sent to the Senate.
Some bills sent to the Senate are not heavily contested -- of the 12 bills that passed in the House, only four had non-trivial action\footnote{``Non-trivial" action is any action that required the votes of individual Senators to be recorded.} once in the Senate.
We focus the remainder of our analysis on these four bills: 117-HR-1319, 117-HR-4346, 118-HR-7888, and 118-HR-3746.

There is typically a sizable period of time between when a bill passes in the House and when the Senate votes.
For the four bills under consideration, there were 10 days, 364 days, 5 days, and 1 day between when the two chambers voted, respectively.
During this time, stakeholders (activists, constituents, lobbyists, etc.) have the opportunity to contact the offices of the Senators to attempt to influence how they will vote.

\begin{figure}
    \centering
    \includegraphics[width=\linewidth]{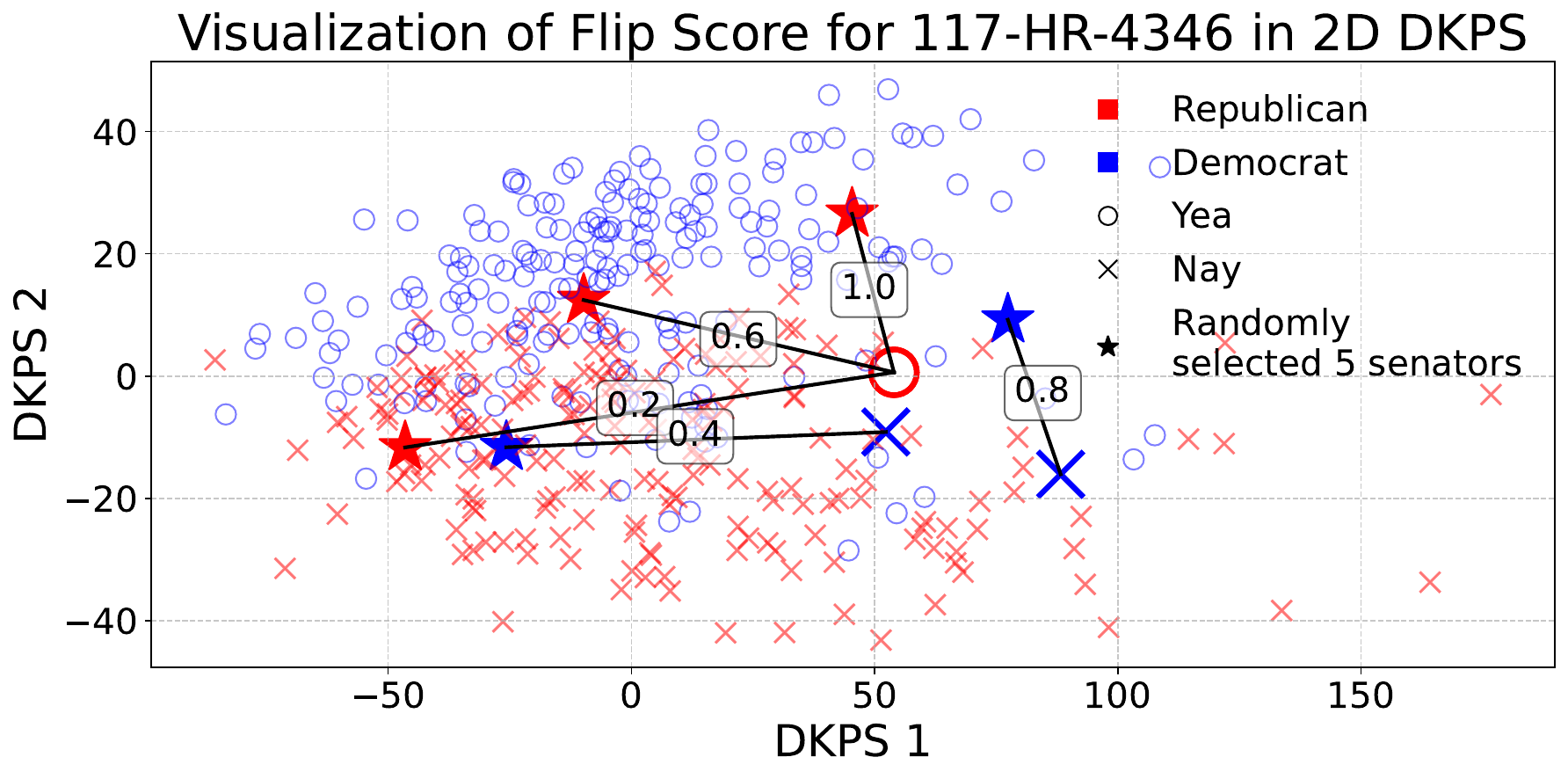}
    \caption{Visualization of the proposed ``flip score" for 117-HR-4346 in 2-d DKPS. 
    Each marker represents a congressperson.
    Color corresponds to party affiliation.
    Representatives who voted ``Yea" are circles ($\circ$) and Representatives who vote ``Nay" are as ``x"es ($\times $). 
    A star ($\star$) represents a Senator whose vote is unknown. 
    Enlarged $\times$ and $\circ$ symbols represent the House member used to calculate each senator's flip score. 
    The flip score for each senator is provided on the line connecting them to the nearest cross-party line voter in their party.
    Senators closer to a cross-party line voter in their party are assigned a higher flip score.
    }
\label{fig:flip-score-dkps}
\end{figure}

Without additional information, the default prediction for how a senator will vote is along party lines. 
For example, if the majority of the senator's fellow party members in the House vote ``Yea" then the senator is likely to vote ``Yea".
As shown in Figure \ref{fig:inference}, when predicting along party lines is accurate, classification using a bill-specific DKPS achieves comparable performance. 
Importantly, when predicting along party lines is not accurate, classification using DKPS is better. 
Thus, in situations where a bill has received votes -- such as when a bill passes or fails in the House -- the geometry of the DKPS can be used to quantify how likely a congressperson is to cross-party lines (or ``flip").
For example, if the DKPS representation of a Republican Senator whose vote is unknown is near the DKPS representation of a Republican Representative who crossed party-line, then they are more likely to cross the party line.
Conversely, if the DKPS representation of the Republican Senator is far away from all Republicans in the House that crossed party lines then they are not likely to cross the party line.


We introduce the ``flip score" to quantify this idea.
For a given bill we identify the closest same-party member of the House that crossed party line for each Senator in the appropriate DKPS.
If there are no cross-party voters in the Senator's party then they are assigned a flip score of $ 0 $.
When there are cross-party voters in the Senator's party then the Senator is assigned a flip score inversely proportional to the distance (in the DKPS) to the nearest same party cross-party voter.
For a given Senator $ S $ and defining 
\begin{align*}
    \mathcal{H}(S) := \{&H: H \in \text{House}, \\ 
    &H \text{ and } S \text{ in same party},\\ 
    &H \text{ voted across party lines}\}
\end{align*}  then, if $ \mathcal{H}(S) $ is non-empty,
\begin{align*}
    \text{flip score}(S) = \frac{1}{\min_{H \in \mathcal{H}(S)} \| X_{H} - X_{S}\|},
\end{align*}  where $ X_{C} \in \mathbb{R}^{d} $ is the DKPS representation of congressperson $ C $.

Figure \ref{fig:flip-score-dkps} shows the DKPS corresponding to 117-HR-4346. 
The figure includes all members of the House who had at least one Tweet before the time of the vote and who voted either ``Yea" or ``Nay".
It also includes five Senators and lines connecting them to the member of the House used to calculate their flip score.

We validate the proposed flip score by comparing it to the observed proportion of Senators who flipped with a given score.
For this, we quantize the flip scores within a given bill.
Flip scores of 0 are assigned a quantized flip score of 0. 
Flip scores in the 0th-20th percentile for the bill are assigned a quantized flip score of 0.2, flip scores in the 20th-40th percentile for the bill are assigned a quantized flip score of 0.4, etc.

\begin{figure}[t]
    \centering
    \includegraphics[width=\linewidth]{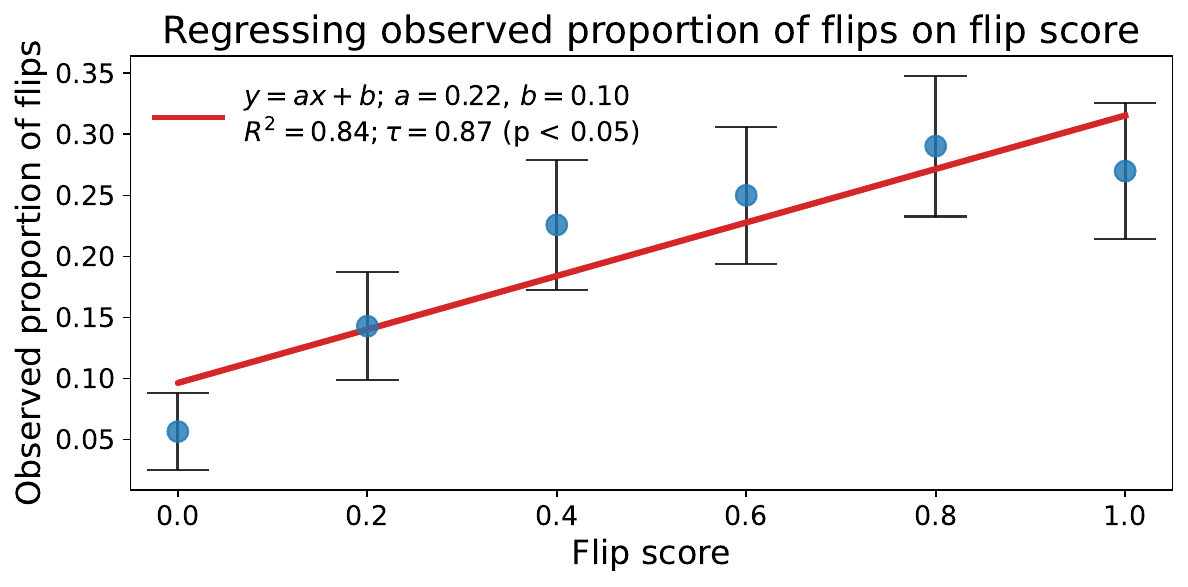}
    \caption{Empirical likelihood of a Senator crossing party-line is correlated with the proposed flip score over four bi-cameral bills. 
    Estimated proportion of flips is calculated by first quantizing the flip scores into 6 bins: one for zero and the other five based on 20\% quantiles (i.e., 0-20\%, 20-40\% , etc.) after removing zeroes.
    Error bars represent one standard error.
    The red line is the linear best fit. The $ R^{2} $ value indicates a strong linear relationship and the p-value corresponding to Kendall's $ \tau $ indicates a statistically significant ordinal association.}
\label{fig:flip-score-validation}
\end{figure}

Figure \ref{fig:flip-score-validation} shows the relationship between the quantized flip score and the observed proportion of flips.
The observed proportion of flips was calculated across all four bi-cameral bills.
As determined by the p-value from the hypothesis test of no ordinal association between the flip score and the observed proportion of flips via Kendall's $ \tau $, there is statistical significance in the relationship (p-value $< 0.05$).
Further, the relationship is quite linear: the linear goodness-of-fit measure $ R^{2} $ is $ > 0.8 $.

Thus, stakeholders can use the flipscores to prioritize how to spend their communication resources: spend more on Senators with a high flip score.
If stakeholders use the flip score to prioritize their communication resources then they provide a mechanism for which the inferences on the virtual models can provide feedback to the physical twin and realize value.
\section{Discussion \& Limitations}
We have provided evidence that a collection of language models each equipped with a congressperson-specific dataset satisfies the four requirements for a virtual model to be a digital twin for a collection of congresspersons.
Indeed, the results herein demonstrate the ability to reasonably implement a system like the one illustrated in Figure \ref{fig:digital-twin}.
While the collection of virtual models satisfies the definition of a digital twin, there are details and limitations of our work that warrant discussion and additional investigation. 

For example, our analysis is focused entirely on Twitter data.
Twitter is by no means the only medium in which congresspersons communicate with the public or each other.
Other sources of data such as campaign speeches, emails to constituents, C-SPAN transcripts, previous voting records, etc. should be included and analyzed before claiming that a collection of virtual models is a proper digital twin -- though there will likely always be a trade-off between virtual model fidelity and virtual model interpretability.

Along the same lines, the data that we use to study the collection of virtual models does not include information required to sufficiently model congressperson-to-congressperson or congressperson-to-public interactions. 
As such, we do not attempt to simulate conversation or approximate more complicated behavior such as drafting a bill.
A virtual model that sufficiently captures these behaviors may be necessary to make a claim of a digital twin for some congressional processes.

We note that previous work has demonstrated the ability to use information about a bill to predict how congresspersons will vote \cite{kraft2016embedding,patil2019roll}.
For example \cite{patil2019roll} trained a classifier to predict voting behavior using a manually curated knowledge base based on news articles by particular congresspersons. Their classifier's accuracy achieves 0.9 on average for all bills voted in 106-109th Congressional sessions. 
Our DKPS-based method performs comparably. Importantly, the DKPS-based method uses the outputs from virtual models of each congresspersons that is demonstrably similar to the outputs of the physical congresspersons.

Figure \ref{fig:inference} compares the utility of the geometry of the retrieved Tweets to the geometry of the generated Tweets.
The geometry of the generated Tweets is more useful than the geometry of the retrieved Tweets for predicting how a congressperson will vote on a particular bill.
This performance gap indicates that \texttt{LLaMa-3-8B-Instruct} is effective at using the retrieved Tweets (that contain little vote-relevant information, per the accuracy) to produce more vote-relevant information.
We note that the base model that we used was released by Meta in mid-April of 2024 and that the majority of the bills we considered were voted on during or after mid-April of 2024.
Hence, we do not expect the ability of the model to produce highly relevant content given low-signal Tweets to deteriorate much as the bills of interest get further away from the time the base model was trained. 

Lastly, we argued that the proposed flip score can be used by stakeholders to improve resource allocation.
As mentioned above, flip score requires existing voting information, such as from the House.
We suspect that unsupervised analogues to flip score will also be useful when optimizing resource distribution.
For example, if the DKPS representation of a Republican is surrounded by DKPS representations of Democrats then there is reason to believe that the Republican may be prone to voting with the Democrats.

Overall, we have provided empirical and statistical evidence that a collection of language models with access to individualized databases that contain Tweets
from the official accounts of congresspersons goes beyond generic “human-like”
generation, behavior, and sociology and reasonably satisfies the definition of a
digital twin.

\subsection*{Acknowledgements} We'd like to thank Henry Farrell, Hahrie Han, Ben Johnson, Connor Pollak, and Jeremy Ratcliff for helpful discussions and feedback throughout the development of this manuscript.
H.M. was supported by the Johns Hopkins SNF Agora Institute.

\bibliography{biblio}

\end{document}